\documentclass[10pt,twocolumn,letterpaper]{article}
\usepackage[utf8]{inputenc} 
\usepackage[T1]{fontenc}    
\usepackage{hyperref}       
\usepackage{url}            
\usepackage{booktabs}       
\usepackage{amsfonts}       
\usepackage{nicefrac}       
\usepackage{microtype}      
\usepackage{algorithm}
\usepackage{algorithmic}
\usepackage{graphicx}
\usepackage{mathtools}
\usepackage{titling}
\usepackage{float}
\usepackage{amsmath,amssymb}
\usepackage{subcaption}
\usepackage{wacv}
\usepackage{times}
\usepackage{epsfig}
\usepackage[symbol]{footmisc}

\DeclareMathOperator{\E}{\mathbb{E}}

\wacvfinalcopy 

\begin{document}

\title{High Fidelity Semantic Shape Completion for Point Clouds using Latent Optimization}

\author{ 
Shubham Agrawal\thanks{Both authors contributed equally}  \\
Carnegie Mellon University\\
{\tt\small shubhamag@cmu.edu}
\and
Swaminathan Gurumurthy* \\
Carnegie Mellon University\\
{\tt\small swamig@cmu.edu}
}

\maketitle

\begin{abstract}
Semantic shape completion is a challenging problem in 3D computer vision where the task is to generate a complete 3D shape using a partial 3D shape as input.
We propose a learning-based approach to complete incomplete 3D shapes through generative modeling and latent manifold optimization. Our algorithm works directly on point clouds. We use an autoencoder and a GAN to learn a distribution of embeddings for point clouds of object classes. An input point cloud with missing regions is first encoded to a feature vector. The representations learnt by the GAN are then used to find the best latent vector on the manifold using a combined optimization that finds a vector in the manifold of plausible vectors that is close to the original input (both in the feature space and the output space of the decoder). Experiments show that our algorithm is capable of successfully reconstructing point clouds with large missing regions with very high fidelity without having to rely on exemplar based database retrieval. 
\end{abstract}
\section{Introduction}

With the increasing availability of low-cost RGB-D scanners, the availability and consequently the need to process 3D data is becoming of great interest to the robotics and vision community. Voxelized representations of 3D data have been quite popular in the learning community because of the ease of generalizing convolution operations to 3D. However most 3D data, whether acquired through RGB-D scanners like Kinect, or through Structure from Motion (SfM) and stereo cameras, is in the form of point clouds. This, along with the fact that point clouds are highly memory efficient while preserving fine surface details, makes it highly desirable to extend deep-learning methods to point clouds. 
Point clouds have been significantly harder to incorporate into deep learning architectures due to irregular organization of points, i.e, they are not regular structures and can't be directly used with architectures that exploit regularity in the input for weight sharing. The networks proposed for point clouds need to be able to handle arbitrarily sized inputs and permutation invariance.\\
A common challenge when reconstructing 3D scenes is that the resulting point clouds may have large missing regions. Reconstructions using SfM may be sparse due to lack of feature points to track on the object. Similarly the point cloud generated by a range scanner may have gaps due to occlusions, limited viewing angle, and may be limited by the resolution of the sensor.
\begin{figure}[H]
\begin{center}
   \includegraphics[width=1\linewidth]{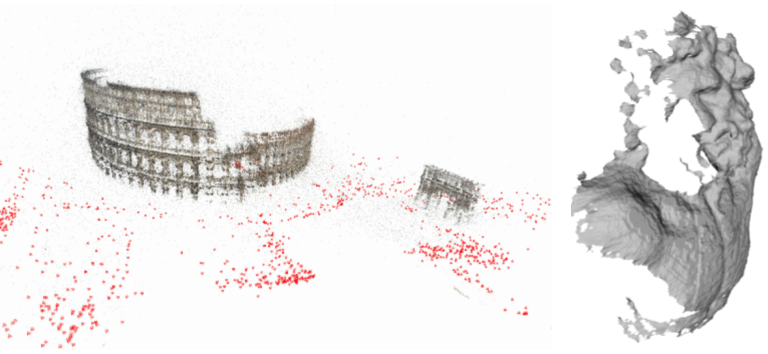}
\end{center}
   \caption{3D reconstruction through techniques like Structure from Motion or RGB-D scanners often leads to incomplete shapes due to lack of feature points and occlusions respectively.}
\label{fig:rome}
\end{figure}

In this paper we aim to solve this challenge using a deep learning approach. We propose a framework that can take as input a point cloud with arbitrary corruption, such as large holes, entire missing regions (such as due to occlusions/viewing angle) and low resolution/ small number of points (which can also be caused by texture-less surfaces during SfM, or due to limitations in the resolution of a RGB-D sensor); and output a dense complete point cloud. Note that the novelty of our approach lies in the fact that we don't need to train on a dataset containing these corruptions and yet manage to handle them at test time. 

Our main contributions are as follows :
\begin{itemize}
    \item The first shape completion framework that works directly on point clouds, and can handle all types of point cloud noises at test time such as large holes, multiple smaller missing regions, and low density, even if trained only on complete point clouds.
    \item A novel algorithm that performs shape completion by performing optimization on a latent manifold learnt by a generative model, using a combination of losses that ensures reconstruction of a valid object while simultaneously fitting the available data.
    \item Quantitative and qualitative evaluation of our method and other baseline methods on both synthetic and real (SfM) data. We demonstrate our method is able to generalize to real data while being trained entirely on synthetic data, something that the baseline methods fail at.
    
\end{itemize}
\section{Related Work}
\textbf{Deep Learning on 3D data.} Common tasks on point clouds include classification, segmentation, object detection and dense labeling \cite{segment,segment2,semantic3d,labeling,sonet}. 
Incorporating point clouds into a deep-learning framework poses several challenges, due to several peculiarities such as input size and order variance, non-uniform density, and shape and scaling variance. \\
Previously, most deep-learning approaches for point cloud centric applications overcame these challenges by voxelizing the point clouds, which allows for the extension of ideas from 2D CNNs into the 3D space \cite{voxnet} \cite{vol_multiview}. However voxels are highly inefficient in terms of accuracy and fidelity of the shape represented, and the network size increases rapidly as spatial resolution is increased \cite{octree}. More importantly, point clouds are the most common and general representation for 3D data as other representations can easily be obtained from them. 

 Qi \etal \cite{pnet,pnetplus}first introduced a deep learning network, PointNet, for point cloud classification and segmentation. The network handles the arbitrary input size of point clouds by using an element wise symmetric operation, such as max pool, to encode any input into a fixed size feature vector.

Achlioptas \etal \cite{latentgan} proposed coupling a PointNet-style encoder with a decoder of fully connected layers, along with loss metric like Earth Mover's distance (EMD) to learn representations of point clouds. They further showed that Gaussian Mixture Models (GMMs) or Generative Adverserial Networks (GANs) could be trained to directly generate the latent representations, which can be used for point cloud generation.

Yu \etal \cite{punet} adapt \cite{pnetplus} for the task of upsampling point clouds. As noted by the authors in the paper, their approach is not suited for point clouds with large gaps or missing regions.

\textbf{Shape Completion.} Shape completion has long been a problem on interest in the graphics and vision community. Traditional geometric methods such as \cite{least_mesh,laplacian,poisson} can only fill in small holes in surfaces. A lot of classical approaches relied on exemplar-based completion, where a CAD database was used to fetch similar models to reconstruct the object, which may then be deformed to match the partial input \cite{shape_single_rgbd,data_driven,retrieval}. 
The vast majority of deep learning works on shape completion have relied on voxel representations due to the ease of generalizing convolution operations to 3D using 3DCNNs. Dai \etal \cite{scancomplete} used 3DCNNs to predict a course complete shape, which was in turn used to lookup similar model from a database. These similar models were then used together with the input for a combined complete shape synthesis. Other recent methods have removed reliance on a database by directly building predictive models for the complete 3D shape, often in a course-to-fine manner \cite{scancomplete,robotic,stutz,highresolution}. Another interesting angle has been the task of predicting 3D shapes from depth maps \cite{pixels,depthviews}, although these don't address the challenge of having point clouds with uneven density and arbitrary holes.

The idea of using deep generative models has been shown to be effective in recovering missing regions in 2D images \cite{inpainting} \cite{glob_local_img_comp}. Similar methods were extended to voxelized 3D shapes recently in \cite{3dinpainting}, which incorporates a convolutional encoder-decoder, a GAN and an LSTM to better learn global and local structure.\\
There has been very little work done on learning-based methods for shape completion directly on point clouds. The authors of \cite{latentgan} show that their autoencoder architecture may also be trained for completing point clouds. More robust formulations of the autoencoder architecture have been proposed in \cite{foldingnet,neighbours} although these works don't address shape completion. We note that these recently proposed approaches aimed at learning more robust representations of point clouds can be seamlessly incorporated into our algorithm, as our latent-optimization is agnostic of the style of encoding-decoding mechanism used.
To the best of our knowledge, we are the first to propose a deep learning based shape completion approach that works directly on point clouds and can handle arbitrary corruptions in the point cloud without requiring any special training. We also need very small networks with much fewer parameters as compared to voxel based approaches. Our approach is purely learning based and does not rely on exemplar-based retrieval from a database, and generalizes well to objects not seen during training.
\section{Methods}
\subsection{Generative models for point clouds}
 We build upon a recent model for point cloud generation proposed by Achlioptas \etal \cite{latentgan}, which extends \cite{pnet} to learn an autoencoder for point clouds. The encoder consists of multiple layers of 1D convolutions followed by a symmetric pooling layer (max pool in our case) resulting in a single global feature vector (GFV) for the entire point cloud. The decoder consists of a set of stacked fully connected layers. The last layer of the decoder outputs an $N\times3$ dimensional vector which corresponds to the N points of the point cloud. The Earth Mover's distance (EMD), which is a permutation invariant metric, is used as the loss function. The EMD between two point clouds $S_1$, $S_2$ is given by:
\begin{equation}\label{eq:EMD}
    d_{EMD}(S_1, S_2) = \min_{\phi:S_1 \rightarrow S_2} \sum_{x \in S_1} {|| x - \phi(x)||_2}
\end{equation}
where $\phi:S_1 \rightarrow S_2$ is a bijection. The optimal bijection is unique and invariant under infinitesimal movement of the points.
The autoencoder is trained on the ground truth complete point clouds in the training set. The trained encoder (E) is then used to extract the global feature vector encoding for each point cloud in the training set. As in \cite{latentgan}, we then train a GAN on the extracted global feature vectors. New feature vectors generated from the generator (G) can be passed through the decoder (H) to generate point clouds. The GAN has the advantage of being a differentiable network - it is possible to take gradients through it from the output to the input distribution space. This is key for our latent-space optimization algorithm.
\begin{figure}
\begin{center}
\includegraphics[width=1\linewidth]{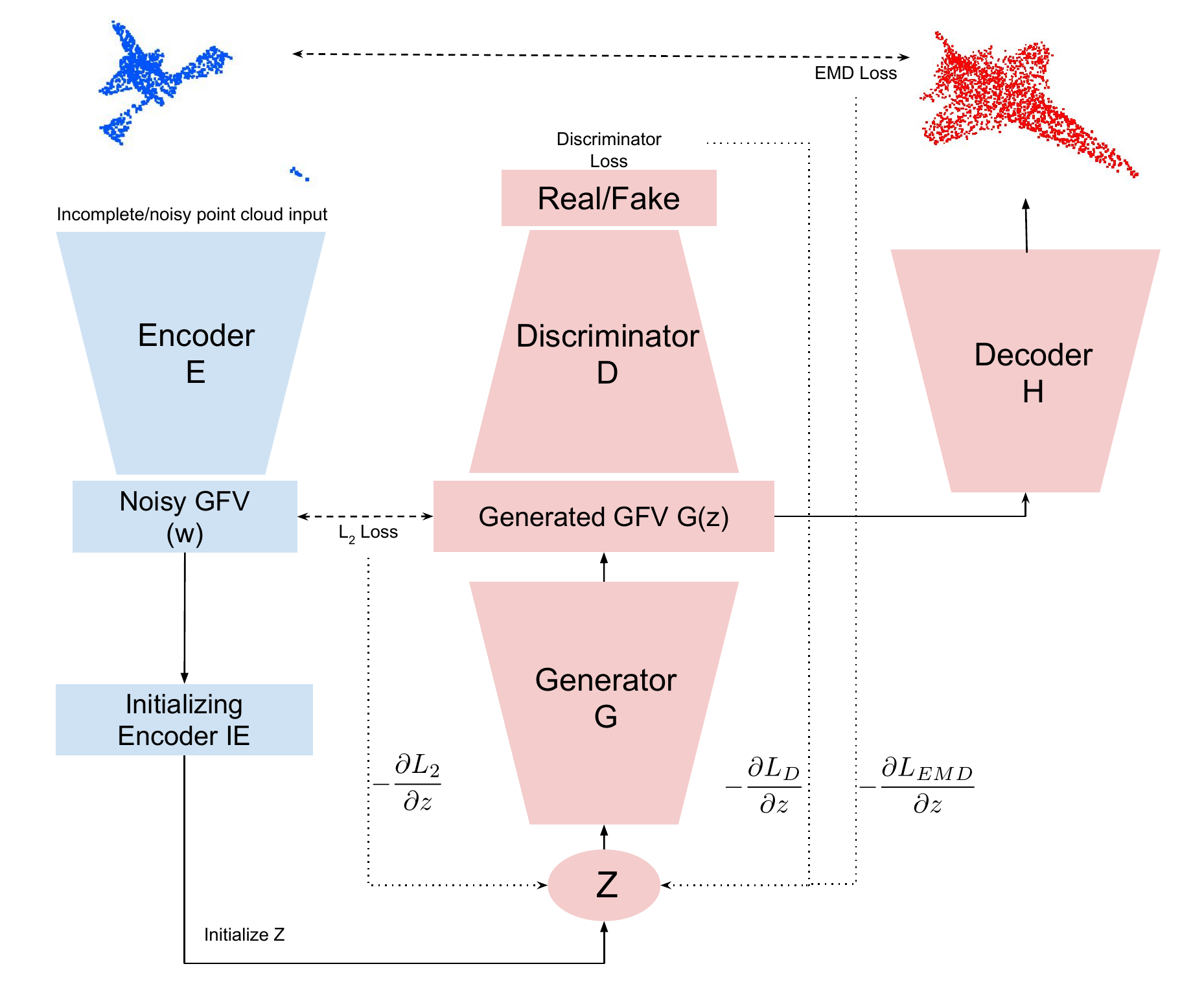}
\end{center}
   \caption{The proposed Point cloud completion algorithm. Loss terms of the LDO for an incomplete input point cloud are visualized with dotted lines. Blocks in blue are only used once for initialization. The losses are used to find the correct point in the latent space of the generator. No network weights are changed.}
\label{fig:ae}
\end{figure}

\paragraph{Generative Adversarial Network (GAN).}
GANs are a popular category of generative models which have been recently shown to produce state of the art results in image generation. GANs learn a mapping from an easy to sample distribution (say, a unit normal distribution) to the data generating distribution using a function approximator like a neural network (generator). The generator(G) is trained in a game theoretic set up, where the objective of the generator is to generate samples that look indistinguishable (to another network, called the discriminator) from the data. The discriminator (D) is trained to distinguish between the real data and the samples generated by G. We introduce a third network, the Initializing Encoder (IE), that learns a mapping from the output of the generator to the latent space $z$. But the traditional GAN training scheme is known to be unstable. Recent advances in GANs \cite{wgan,gan,gan_num} have tried to address this issue by modifying the loss or the training procedure itself. We use the loss modification proposed in \cite{wgan} for more stable training. We train the GAN on the set of global feature vectors(GFVs) produced by the encoder. We use fully connected layers for both the generator and the discriminator. The training procedure for the AE and GAN is given in Algorithm \ref{alg:training}. The losses optimized for training the GAN along side the IE have been described below :

\begin{align}
    J^{(D)} &= \E_{z\sim p_z}{D(G(z))} - \E_{x\sim p_{data}}{D(x)}\\
\begin{split}
 J^{(G)} &= \E_{z\sim p_z}{[||IE(G(z)) - z||_2 - D(G(z)) ]} \\
    & \ \ \ \ + \E_{x\sim p_{data}}{||G(IE(s)) - x||_2} 
\end{split}\\
\begin{split}
    J^{(IE)} &= \E_{z\sim p_z}{||IE(G(z)) - z||_2} \\
              & \ \ \ \ + \E_{x\sim p_{data}}{||G(IE(s)) - x||_2}
\end{split}
\end{align}
where, $p_z$ is the unit normal distribution centered at the origin, $p_{data}$ is the distribution of GFVs, $z$ and $x$ are samples from these distributions.
\subsection{Point Cloud Completion using LDO}
 \label{sec:ldo}
Consider an incomplete point cloud at test time, such as one generated via SfM. The point cloud may also be noisy and have uneven density. If this point cloud is passed through the encoder E, a "noisy" GFV is obtained, i.e. one that doesn't lie on the manifold of representations learnt by the autoencoder. We model the task of completing the point cloud as obtaining a clean GFV corresponding to the noisy one, through an optimization procedure. The cleaned GFV can then be passed through the decoder (H) to obtain a completed point cloud. 

Thus the task is reduced to projecting the noisy GFV onto the manifold of clean GFVs. This is not trivial, since we don't have an analytical expression to represent the clean GFV manifold. Thus we use a GAN to represent the clean GFV manifold. As described in the previous section the GAN is trained on clean GFVs, extracted from the training set of complete point clouds. Projecting a noisy data point onto the manifold of clean GFVs can be reduced to finding the closest GFV to the noisy GFV, that is also classified as real by the discriminator. However, directly optimizing over the space of GFVs would result in adversarial examples. Thus we choose to perform the optimization procedure in the latent space of the generator, represented by the latent vector $z$. First we produce an initialization for $z$ by passing the noisy GFV through the Initializing Encoder, $z_{init} = IE(GFV)$. From this initial value, $z$ is optimized so as to produce a clean GFV through the generator, $G(z)$. Specifically, the objective of our Latent Denoising Optimization (LDO) algorithm can be decomposed into three parts:

\textbf{Discriminator Loss:} This term ensures that the generated GFV is from the data manifold. We optimize to maximize the score given by the discriminator to the generated GFV:
\begin{equation}
    L_D(z) = - D(G(z))
\end{equation}

\textbf{Latent Least Squares Loss:} This term ensures that the generated GFV, $G(z)$ is close to the noisy GFV, $w_i$ during the optimization and thus remains semantically similar to the input point cloud. The noisy GFV was obtained using the encoder E, $w_i = E(S_i)$. We simply minimize the L2 distance between the generated GFV and the noisy GFV:
\begin{equation}
    L_2(z;w) = ||G(z) - w_i||_2^2
\end{equation}

\textbf{Decoder EMD Loss:} This term ensures that the generated GFV maps to a point cloud which is close to the input point cloud where it exists. Here, we minimize the Earth Mover's distance between the input point cloud and the point cloud decoded from the generated GFV:
\begin{equation}
    L_{EMD}(z;S_i) = d_{EMD}(S_i, H(G(z)))
\end{equation}


Thus our final loss becomes a weighted combination of these losses:
\begin{equation}
    Loss(z) = L_{EMD}(z;S_i) + \lambda L_D(z) + \beta L_2(z;w_i) 
\end{equation}\\
We perform this optimization using the ADAM optimizer. Note that we use an exponential decay to reduce the value of $\lambda$ and $\beta$, starting with an inital value of $0.001$ each. This helps us ensure that in the initial stages of the optimization, the emphasis on obtaining a semantically consistent and real looking point cloud and in the latter stages, the emphasis is on reconstructing fine details of the point cloud. The optimization is stopped as soon as the loss $L_D(z)$ starts increasing. This ensures that the optimization does not lead to 'unreal' looking point clouds in order to get the details right. It is important to note here that we only minimize the loss with respect to $z$ (which is the input to the Generator). We do not update the generator or discriminator parameters when performing this optimization. At the end of the optimization, we obtain the optimal latent vector, $z^*$. We simply pass this through the generator to get the clean GFV, $G(z)$, which is passed through the decoder to obtain the completed point cloud, $H(G(z))$. The entire Latent Denoising Optimization (LDO) algorithm has been given in Algorithm \ref{alg:ldo} and visualized in fig \ref{fig:ae}. Note that the hyperparameter values stay the same for all our experiments. Thus no experiment specific tuning is required.

 \begin{algorithm}[H]
 \caption{Training a generative model for use in LDO algorithm}\label{alg:training}
 \begin{algorithmic}[1]
 \renewcommand{\algorithmicrequire}{\textbf{Require:}}
 \renewcommand{\algorithmicrequire}{\textbf{Require:}}
 \REQUIRE A training set of clean, complete point clouds S.
 \STATE Train an autoencoder, with encoder E and decoder H, on the training set S, using EMD as the loss metric. For our experiments we use the implementation in \cite{latentgan}, but our algorithm is completely transferable to other PointNet-style architectures such those presented in \cite{pnetplus,foldingnet}.
\STATE Using the trained Encoder, extract the global feature vectors (GFVs) for all examples in the training set.
\STATE Train a GAN to fit on the distribution of extracted GFVs from training set.
 \end{algorithmic} 
 \end{algorithm}

 \begin{algorithm}[H]
 \caption{Point cloud completion using LDO algorithm}\label{alg:ldo}
 \begin{algorithmic}[1]
 \renewcommand{\algorithmicrequire}{\textbf{Require:}}
 \renewcommand{\algorithmicrequire}{\textbf{Require:}}
 \STATE Extract the GFV $w_i$ for the partial cloud $S_i$ by passing it through Encoder E. $w_i = E(S_i)$\\
 \STATE Initialize the latent vector $z_i$ using the initializing encoder IE, $z_i = IE(w_i)$\\
 \STATE Set $ prevLD = L_D(z_i), \lambda=0.001, \beta = 0.001$
  \FOR {k=1,2..N}
  \STATE Compute $\nabla_z Loss(z_i) = \nabla_z L_{EMD}(z_i;S_i)+ \lambda \nabla_z L_D(z_i) + \beta \nabla_z L_2(z_i;w_i)$
  \STATE Update $z_i$ using ADAM
  \STATE Update $\lambda = \lambda*0.9998$
  \STATE Update $\beta = \beta*0.9998$
  \IF {$L_D(z_i)>prevLD$}
  \STATE Exit Loop
  \ENDIF
  \STATE Update $prevLD = L_D(z_i)$
  \ENDFOR

  \STATE Pass the cleaned GFV $G(z_i)$ through the previously trained autoencoder's decoder D to obtain the semantically completed point cloud $H(G(z_i))$

 \end{algorithmic} 
 \end{algorithm}

\section{Experimental Results}
In this section we demonstrate the salient features of our LDO algorithm by evaluating its quantitative and qualitative performance in multiple scenarios. We compare our model to 2 baseline models of point cloud completion, and show the improvement in reconstruction by applying LDO in conjunction with these baseline models. We show quantitative and qualitative results of the improvements gained by LDO on the tasks of point cloud completion and upsampling.
Finally, we show experiments on a real world scenario (SfM) where we demonstrate that our approach particularly shines in generalizing to real world data, while only having been trained on synthetic data.
To summarize, we'll be comparing the following models:
\begin{enumerate}
    \item Autoencoder (\textbf{AE}): An autoencoder trained only with complete point clouds. See Appendix for full implementation details. This baseline is used to demonstrate cases where no prior information is available about the deformities in the true data.
    \item Denoising Autoencoder (\textbf{DAE}): Another intuitive baseline is an autoencoder with the same architecture as AE, trained with an augmented dataset of incomplete point clouds.  While working on our experiments, we became aware that the authors of \cite{latentgan} incidentally updated their work to suggest a similar DAE based completion method. Our initial experiments found that DAEs have a tendency to overfit, and don't generalize well to different amounts and kinds (small holes, large missing region, low-resolution, SfM point clouds) of incompletion (see appendix). Hence, we train different DAEs with different amounts of incompletion and test them on the same amounts of incompletion to get a competitive baseline for comparison. Although one would never have this luxury in the real world, this baseline is used to demonstrate the superiority of our method even when the exact kind and amount of deformity is known beforehand. 
    \item Latent Denoising Optimization with AE (\textbf{AE+LDO}): Our algorithm applied using AE and a GAN learnt on GFVs of the clean training data generated by the AE. We show that \textit{despite never having been trained on noisy/incomplete point clouds,} AE+LDO is very effective at point cloud completion and achieves a huge boost over just the AE's performance.
    \item Latent Denoising Optimization with DAE  (\textbf{DAE+LDO}): To show the transferability of LDO, we also apply it on DAE, with a GAN trained on GFVs produced by the DAE on clean training data. We show that LDO is able to capitalize on the more robust representations learnt by DAE to improve performance even further than AE+LDO.  
\end{enumerate}
\textbf{Dataset.} We use ShapeNetCore, a subset of the full ShapeNet\cite{shapenet} dataset with manually verified category and alignment annotations. It covers 55 common object categories with about 51,300 unique 3D models. For the purposes of our experiments we use 4 classes with the most available data from the dataset, namely: airplane, car, chair and table. For each class, we split the models into 85/5/10 train-validation-test sets for our experiments and results. We use the models without any pose or scale augmentations. We uniformly sample the point clouds (2048 points each) from these models, which serve as the ground truth for our training. 
In section 4.3, we experiment on a real-world data case we take sequences of images of faces and pass them through an SfM pipeline to get noisy point clouds of faces. We use the CMU Multi-PIE \cite{multipie} dataset as the source of these face images and the Basel face model \cite{basel} to obtain a synthetic dataset of faces. More details on this are provided in section 4.3

\subsection{Masking Experiments}
\label{sec:masking}
For the first set of experiments we choose a synthetic masking scheme to demonstrate the benefits of using LDO in point cloud completion tasks. In order to perform masking on a point cloud, we first choose a random point from the point cloud, and remove its 2048*(X/100) nearest neighbors of the point to obtain an X\% masking. To ease batch processing of point clouds with unequal number of points, we replicate one of the non-masked points so that each point cloud is the same size. The PointNet architecture by its nature ignores replicated points. In latter sections we would look into more realistic scenarios where this would become important. 
\subsubsection{Varying levels of Incompletion}
We train a vanilla autoencoder (AE) using the training set in the Airplane class. We then train a GAN on the GFVs of the AE. At test time, we test the AE and AE + LDO (Latent Denoising Optimization) with varying levels of masking ($20\%, 30\%, 40\%$ and $50\%$) in the input point cloud. The corresponding scores have been reported in Table \ref{tab:scores}. Note that we didn't have to retrain our model for the different levels of masking. We observe a clear improvement in performance using our method. We also note that the performance of AE decreases with increasing levels of masking whereas AE+LDO remains more or less robust.
We also train multiple denoising autoencoders (DAE) along with the corresponding GANs with varying levels of masking in the input ($20\%, 30\%, 40\%$ and $50\%$). The DAEs are trained to reconstruct the ground truth given the masked input. In this case, we test DAE and DAE + LDO with masking amounts that the DAE and the corresponding GAN was trained on. So a DAE and GAN trained with $40\%$ masking are tested on $40\%$ masking. This is done to demonstrate the benefit of LDO even when the underlying model (DAE) already has prior knowledge about the incompletion (since it was trained on the specific kind incompletion). The corresponding scores have been reported in Table \ref{tab:scores}. We observe that AE + LDO perform on par with a DAE despite not having any prior knowledge about the kind or amount of incompletion during training. Moreover, performing LDO on DAE provides further improvement as seen from the scores in Table \ref{tab:scores}. This shows that our model can integrate with any AE architecture and capitalize on the robust representations learnt by the models. A visualization of how reconstruction quality varies with increasing percentage of missing data can be found in the appendix. 

\subsubsection{Different classes}
\begin{figure}
\begin{center}
\includegraphics[width=1\linewidth]{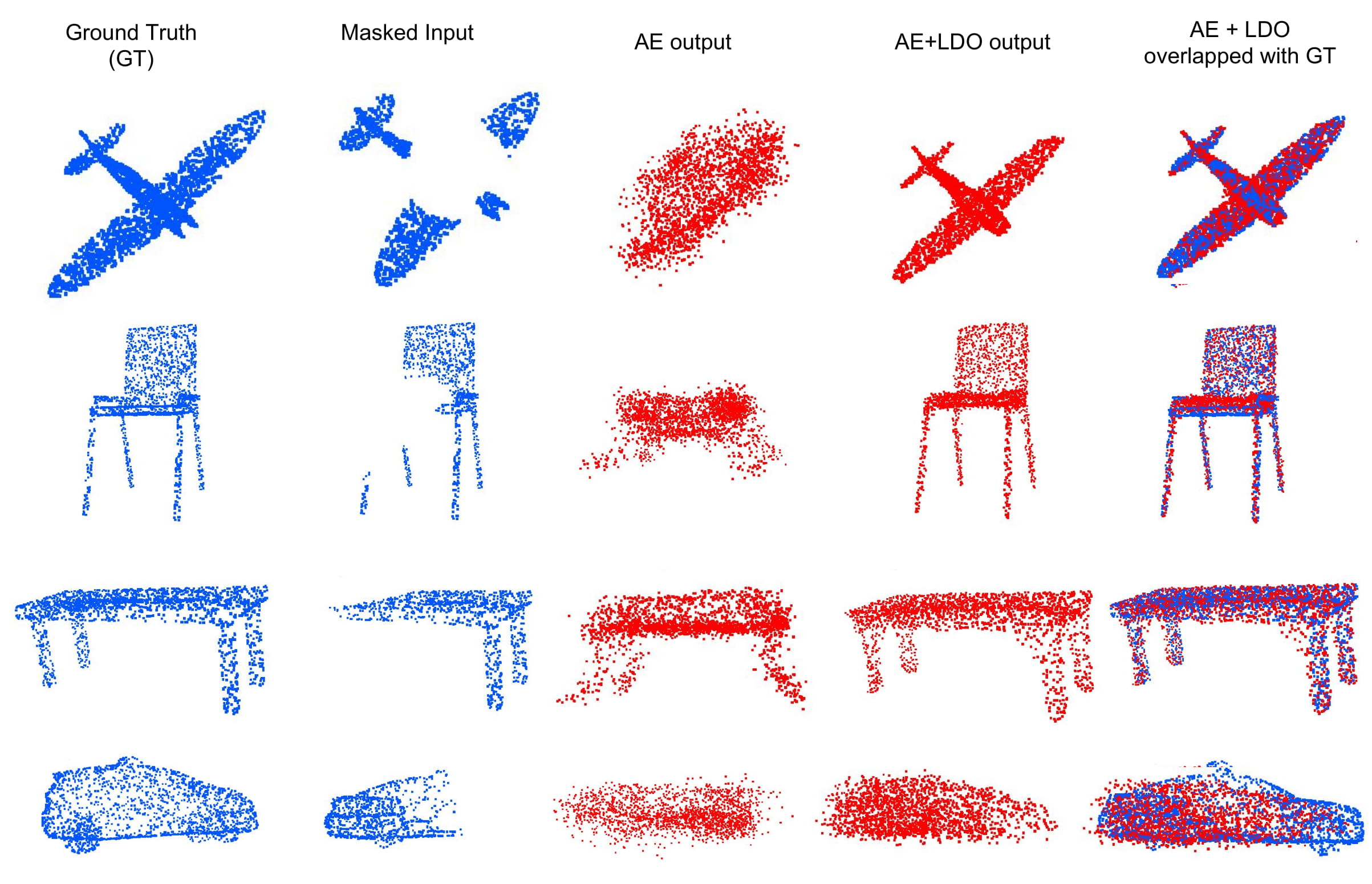}
\end{center}
   \caption{Visualizations of shape completion results of AE and AE+LDO on a test set containing all 4 classes. Random 50\% chunks of the inputs are masked at test time. The outputs under "AE" are from a single autoencoder trained on objects of all 4 classes. The right most column shows the AE+LDO outputs overlapped with the ground truth for direct comparison. A massive improvement is seen in reconstruction quality with our method.}
\label{fig:masked_ae}
\end{figure}

To show the robustness of our methods, we test our model on other classes, namely Chair, Car and Table, both in single-class and multi-class setups. We train a separate AE and the corresponding GAN on the training set of each category. We also train separate DAEs and corresponding GANs with $30\%$ and $50\%$ masking for each category. We also train a multi-class AE, GAN pair on a training set with a combination of the four classes (Table, Chair, Car and Airplane). Correspondingly we train two DAE, GAN pairs with $30\%$ and $50\%$ masking respectively (referred to as Multi-Class in results tables). We test these models on $30\%$ and $50\%$ masking using the corresponding test sets and report the results in Table \ref{tab:scores}. As in the previous section, the DAEs trained with specific masking amounts are tested with the same masking amounts. We observe a similar pattern as was observed in Airplane in all classes except Cars. AE+LDO performs on par or better than DAE in most of these cases. Moreover, incorporating LDO with DAE further improves the results and provides better scores than all other models in most cases. Interestingly, DAE trained on masked Cars performs better or on par with our models. On further inspection we find that this is because there is very little variety in the dataset of cars. Thus DAE is able to easily transfer from the Car training set to the Car test set by simply producing the nearest neighbors from the training set. 
We visualize completion results with 50\% masking using our best performing multi-class model (DAE + LDO) and compare it against its baseline (DAE) in Figure \ref{fig:results}(An enlarged version may be found in the appendix). It is seen that our multi-loss optimization ensures that both, a sharp, valid object is reconstructed, that also fits the available partial scan as best as possible. Figure \ref{fig:masked_ae} compares the point cloud completion results of AE and AE+LDO with 50\% masking. We observe that AE produces meaningless point clouds when the inputs are very highly masked/distorted. Yet, just the addition of our algorithm drastically boosts the quality of results as shown in the figure, despite the models never having been trained on incomplete point clouds.

\begin{figure}
\begin{center}
\includegraphics[width=1\linewidth]{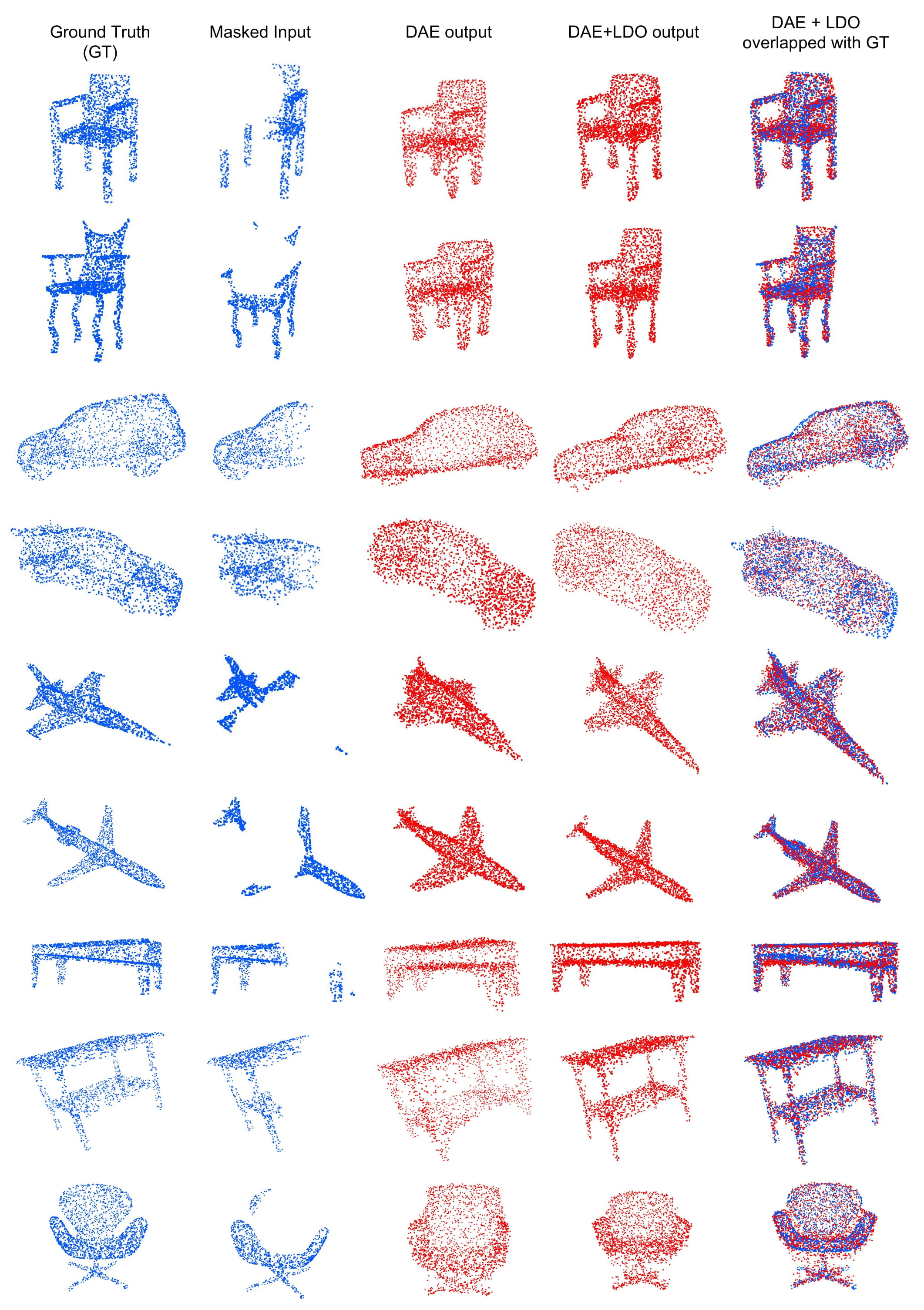}
\end{center}
   \caption{Visualizations of shape completions of LDO on a test set containing all 4 classes. The outputs under "DAE" are from single denoising autoencoder trained on objects of all 4 classes, with 50\% missing data. Outputs of DAE+LDO are of the single DAE and a single GAN trained on global feature vectors of the DAE. DAE+LDO leads to much sharper outputs with more details of the partial shape captured. Last column shows ground truth and our results overlaid for ease of comparison.}
\label{fig:results}
\end{figure}

\begin{figure}
\begin{center}
\includegraphics[width=1\linewidth]{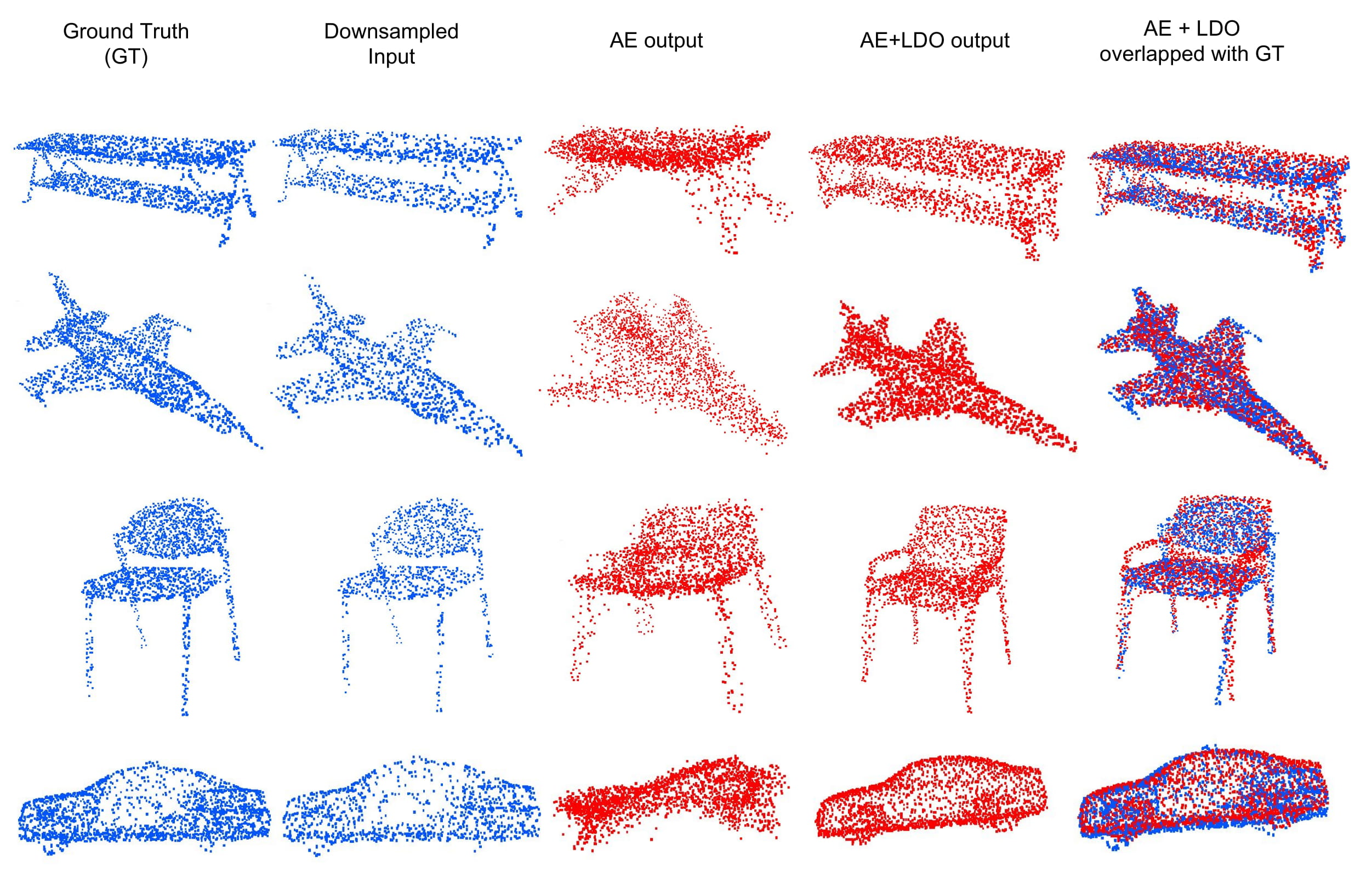}
\end{center}
   \caption{Visualizations of upsampling results of AE and AE+LDO on a test set containing all 4 classes. The inputs at test time are downsampled to 1/5th of the original points. The outputs under "AE" are from a single autoencoder trained on objects of all 4 classes. The right most column shows the AE+LDO outputs overlapped with the ground truth for direct comparison.}
\label{fig:upsampling}
\end{figure}

\begin{figure}
\begin{center}
\includegraphics[width=1\linewidth]{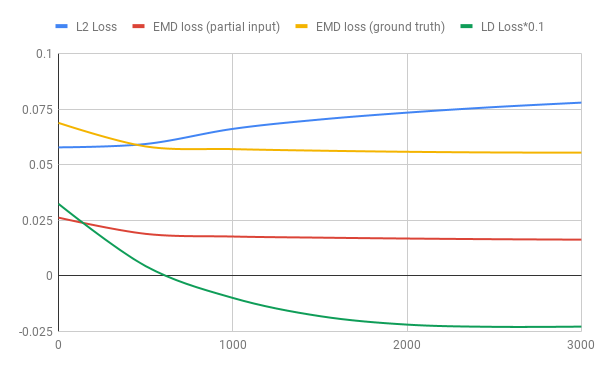}
\end{center}
   \caption{Plot of losses of a typical LDO optimization. EMD loss against ground truth is used for evaluation, not for optimization. LD loss is scaled by 0.1 for ease of visualization}
\label{fig:plot}
\end{figure}

\begin{table}
\begin{center}
\resizebox{\columnwidth}{!}{%
\begin{tabular}{ |c|c|c|c|c|c|} 
 \hline
Category & \% Points Missing & AE & DAE & AE + LDO(ours) & DAE + LDO(ours) \\
\hline
Airplane & 20\% & 0.061 & 0.033 & 0.030 & \textbf{0.028} \\
Airplane & 30\% & 0.079 & 0.036 & 0.037 & \textbf{0.033} \\ 
Airplane & 40\% & 0.083 & 0.039 & 0.041 & \textbf{0.034}\\
Airplane & 50\% & 0.097 & 0.039 & 0.038 & \textbf{0.037} \\
Chair & 30\% & 0.107 & 0.061 & 0.052 & \textbf{0.050} \\
Chair & 50\% & 0.120 & 0.064 & 0.069 & \textbf{0.055} \\
Car & 30\% & 0.096 & 0.0427 & 0.054 & \textbf{0.041} \\
Car & 50\% & 0.118 & \textbf{0.046} & 0.060 & 0.051 \\
Table & 30\% & 0.142 & 0.055 & 0.052 & \textbf{0.047}\\
Table & 50\% & 0.143 & 0.055 & 0.062 & \textbf{0.050} \\
Multi-Class & 30\% & 0.121 & 0.072 & 0.058 & \textbf{0.044} \\
Multi-Class & 50\% & 0.113 & 0.069 & 0.056 & \textbf{0.046} \\
\hline
\end{tabular}
}
\end{center}
\caption {EMD loss of completed point clouds against ground truth (lower is better). As baselines we compare against an autoencoder(AE) trained only with complete point clouds as well as a denoising AE (DAE) trained with partial point clouds. For fairness, the DAEs were trained with the same percentage of incompleteness as they were tested against. We report the performance of our LDO algorithm when used together with the AE (AE + LDO) and with the DAE(DAE+LDO). Multi-Class refers to training a single AE/DAE to reconstruct all 4 classes, as well as our own algorithm when used with these AE/DAEs } \label{tab:scores}
\end{table}

\subsection{Upsampling Experiments}
Upsampling is another important task that comes up in processing 3D data. This is especially important for SfM methods, that often rely on sparse feature points. We investigate the performance of LDO for upsampling point clouds that had been downsampled to 20\% points of the original, using just a regular AE without any special training, and see impressive results. We show the EMD loss for plain AE and our model in Table \ref{tab:upsample}. The upsampled visualizations are given in Figure \ref{fig:upsampling}. We see that the AE struggles to reconstruct any meaningful point clouds. Yet, just by the addition of our algorithm (AE+LDO) we observe a tremendous improvement in the upsampling quality. This shows the versatility of our approach.
\begin{table}
\begin{center}
\begin{tabular}{ |c|p{2.5cm}|c|c|c|c|} 
 \hline
Category & Amount of downsampling at input  & AE & AE + LDO \\
\hline
Multi-Class & 80\% & 0.073 & \textbf{0.058} \\
\hline
\end{tabular}
\end{center}
\caption {EMD loss of upsampled point clouds against ground truth(lower is better). As baselines we compare against a autoencoder(AE) trained only with complete point clouds. We report the performance of our LDO algorithm when used together with the AE (AE + LDO).} \label{tab:upsample}
\end{table} 

\subsection{Real-world Experiments}
\label{sec:rw}
\begin{figure}
\begin{center}
\includegraphics[width=1\linewidth]{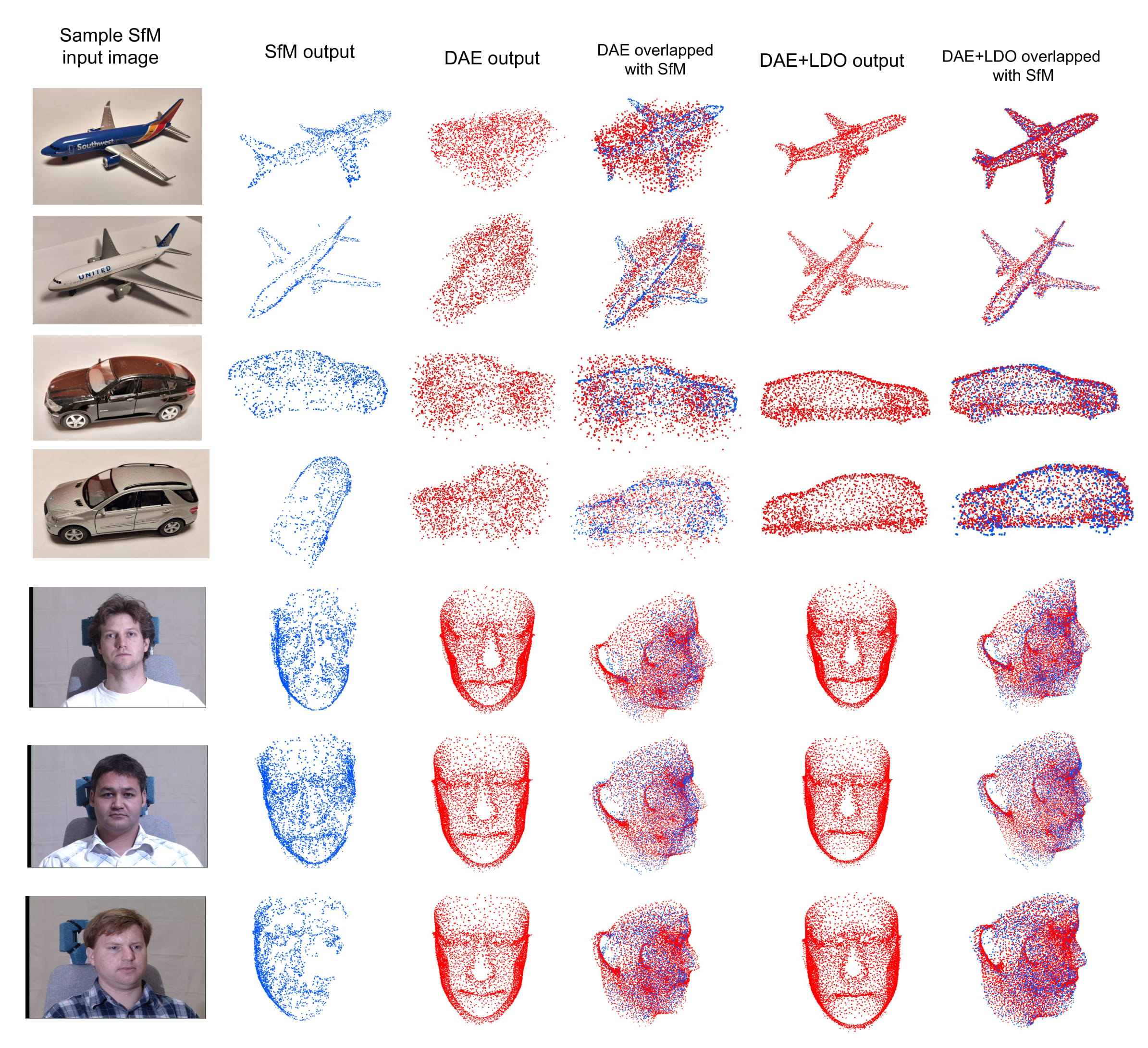}
\end{center}
   \caption{Visualizations of shape completion task on noisy and incomplete point clouds generated by a general-purpose SfM pipeline.}
\label{fig:sfm_ldo}
\end{figure}

To evaluate the real-world applicability of LDO, we test it on the task of completing input point clouds obtained from SfM. The aim is to see whether LDO can generalize to real-world point cloud data, \textit{while having been trained only on synthetic data}. We use COLMAP \cite{colmap}, a general purpose SfM pipeline to generate point clouds using sequences of images.  We experiment with the following classes:
\begin{enumerate}
    \item Shapenet type models: We use some toy car and airplane models for testing. The models were placed on a rotating surface and multiple images were clicked from different poses around the object. These images were then processed through COLMAP. The output point cloud had incompletions due to severe lack of texture on these models. We test these incomplete point clouds on the multi-class DAE and DAE+LDO, trained with 50\% masking on ShapeNet models, as described in section 4.1.2. We choose these, as they were the best performing models in our previous experiments.
    \item Faces : We first create a synthetic training set of 3000 face point clouds using the Basel 3D Morphable Model \cite{basel}. The model provides a PCA basis for faces, and different faces can be obtained by sampling the PCA coeffecients from gaussians. We train a DAE with similar architecture as the ShapeNet DAEs, except with an input/output size of 8192x3. It is trained with 50\% masking on the synthetic face dataset. We then also train a GAN on the GFVs obtained by this DAE, as in the regular procedure to setup LDO. 
    Next, we use the CMU Multi-PIE \cite{multipie} to obtain a sequence of images of human faces taken from different poses. These are processed through COLMAP to obtain point clouds. These are tested on the DAE and DAE+LDO models trained on the synthetic Basel dataset.
\end{enumerate}
For both classes we align and do a "rough" cleaning of the obtained point cloud by aligning it against a template point cloud of the corresponding synthetic set, and removing points beyond a threshold distance from the template. Note that is only to remove background points - the intrinsic noisiness of the points characteristic of SfM is preserved.  Where needed we also downsample them to fit our model resolution. The qualitative results can be seen in Fig \ref{fig:sfm_ldo}. It is observed that the DAE by itself, having only been trained on synthetic data, fails completely on the ShapeNet-type models, and reconstructs badly fitting faces for the face models, since it fails to generalize beyond the PCA-basis constrained synthetic faces. However DAE+LDO gets high quality reconstructions that fit well with the partial input (A larger image visualizing the face reconstructions from profile views is given in the Appendix). 

\subsection{Analysis of loss functions}
We show a plot of the 3 losses used in LDO optimization and the EMD loss against ground truth (used for evaluation) during the optimization process of DAE + LDO in Fig \ref{fig:plot}. The x-axis shows the number of iterations and the y-axis shows the loss values as the optimization progresses. The plot shows that the initialization encoder provides a decent initialization for the optimization, as measured by the ground truth EMD Loss (EMD-GT). This shows that the initialization itself is decent enough to provide scores competitive to that of the DAE. The optimization that follows is responsible for the improvements over the baseline DAE model. We observe that throughout the optimization, the three losses, namely, Partial-EMD Loss, Discriminator loss(LD Loss) and EMD-GT Loss decay gradually until the end of optimization, whereas the L2 loss increases gradually. This indicates that the GFV is being cleaned as it moves away from the noisy GFV and moves closer to the clean GFV. The optimization highlighted here takes 324 sec to process 50 point clouds on a Titan X GPU. The batch size could be increased to achieve lower time per point cloud.

\section{Discussion and Conclusion}

In this work, we presented a novel scheme for point cloud completion using a purely learning based approach. We demonstrate the following salient features of our approach : \begin{itemize}
    \item We show the superiority of our algorithms on a variety of incompletion types ranging from large missing regions (masked), low-density point clouds (upsampling) to real world SfM point clouds.
    \item Even when trained with only complete point clouds, our algorithm (AE+LDO) was able to obtain high quality reconstructions.
    \item Compared with a denoising autoencoder baseline, our approach was shown to generalize much better to unseen data (sec \ref{sec:rw}, \ref{sec:masking}). The reconstructions obtained by DAE+LDO were shown to have higher fidelity to match the partial input, whereas the DAEs overfit to the training data resulting in more generic reconstructions.
    \item LDO shows generalization even in scenarios when the underlying model is trained on synthetic data and then tested on real world data (sec \ref{sec:rw}). In fact, even in cases where the DAE fails completely, DAE+LDO is able to extract high quality reconstructions.

\end{itemize}
LDO is quite flexible to the actual autoencoder architecture and training mechanism used. We show that it is able to capitalize on the more robust representations learnt by a DAE. Recent works have proposed more robust formulations of encoder-decoder architectures for pointclouds, such as by incorporating local neighbourhood information \cite{pnetplus,foldingnet,neighbours}. In our future work we wish to explore the integration of LDO into these frameworks. We would also like to explore better optimization schemes than ADAM.

\section*{Acknowledgments} 
We would like to thank Panos Achlioptas for sharing his implementation of the autoencoder and Haoqiang Fan for the CUDA implementation of EMD loss function.

\newpage

\subsection{Implementation Details}
\begin{figure*}
\begin{center}
\includegraphics[width=1\linewidth]{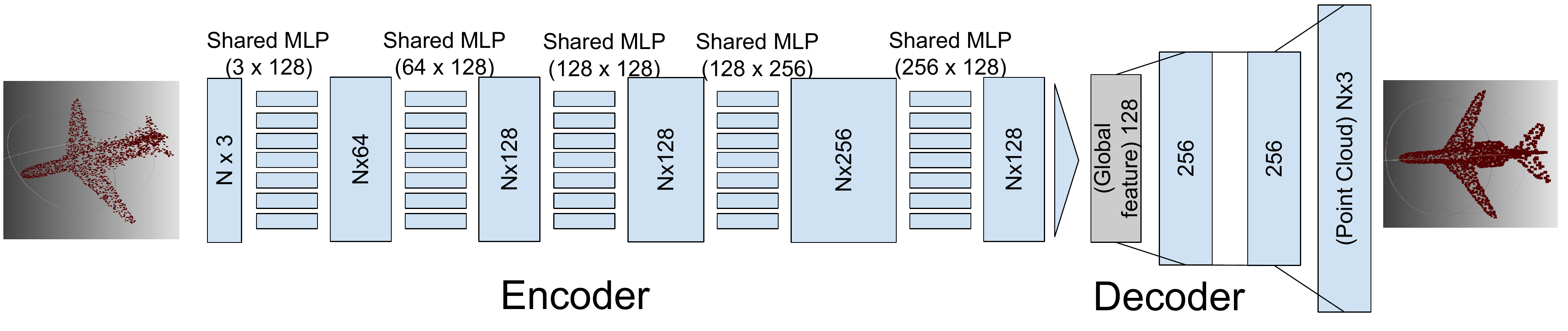}
\end{center}
   \caption{\textbf{Autoencoder Architecture.}}
\label{fig:architecture}
\end{figure*}

We implemented the Autoencoder as proposed by Achlioptas \etal in [5]. The encoder consists of 5 layers of 1x1 convolutions with 64, 128, 128, 256 and 128 filters respectively. Each layer is followed by a ReLU non-linearity. This is followed by a global max pooling operation leading to a bottleneck size of 128. The decoder consists of three fully-connected layers leading to hidden layers of size 256, 256 and 6144 respectively. The output of the last layer is reshaped to 2048$\times$3 to get the output point cloud. We use the same architectures for the Autoencoder and Denoising Autoencoder. Figure \ref{fig:architecture} shows more details about the specific architecture. When feeding an incomplete point cloud we simply remove those points from the input. Since the operations performed are symmetric to all points and agnostic to the number of point clouds, the architecture works as is for incomplete point clouds as well.
The AEs were trained using an optimized approximate implementation of the the EMD loss, as used by Fan \etal in [3]. We also use EMD of the completed point clouds against the respective ground truth to evaluate our models. 
We use this AE to generate a set of global feature vectors for all the point clouds in the training set. We then train a W-GAN to be able to generate samples from this distribution of global feature vectors. The Generator samples from a unimodal gaussian noise distribution of 128 dimensions. It consists of two fully connected layers of size 128 each, and outputs a global feature vector of size 128. The discriminator takes a 128 dimension vector at its input. It has 2 fully connected layers of size 256 and 512 followed by a final layer which outputs one value and a sigmoid activation, resulting in a prediction of whether the input vector was real or fake. The initialization encoder is trained to map each GFV to a corresponding latent vector. It also consists of 2 fully connected layers of size 128 each. It takes the GFV as input and outputs the 128 dimensional latent vector which is taken as input by the Generator and mapped back to the corresponding GFV. The GAN is trained with a learning rate of 0.0001 and an ADAM optimizer, for 200 epochs. We use the same learning rate with ADAM optimizer for LDO as well. Initial values of $\lambda$, $\beta$ and $\alpha$ are 0.1, 0.1, 0.001 respectively. $\lambda$ and $\beta$ are further decayed by a factor of 0.999 after every update. 

\section{DAE overfitting}
\begin{figure}
\begin{center}
\includegraphics[width=1\linewidth]{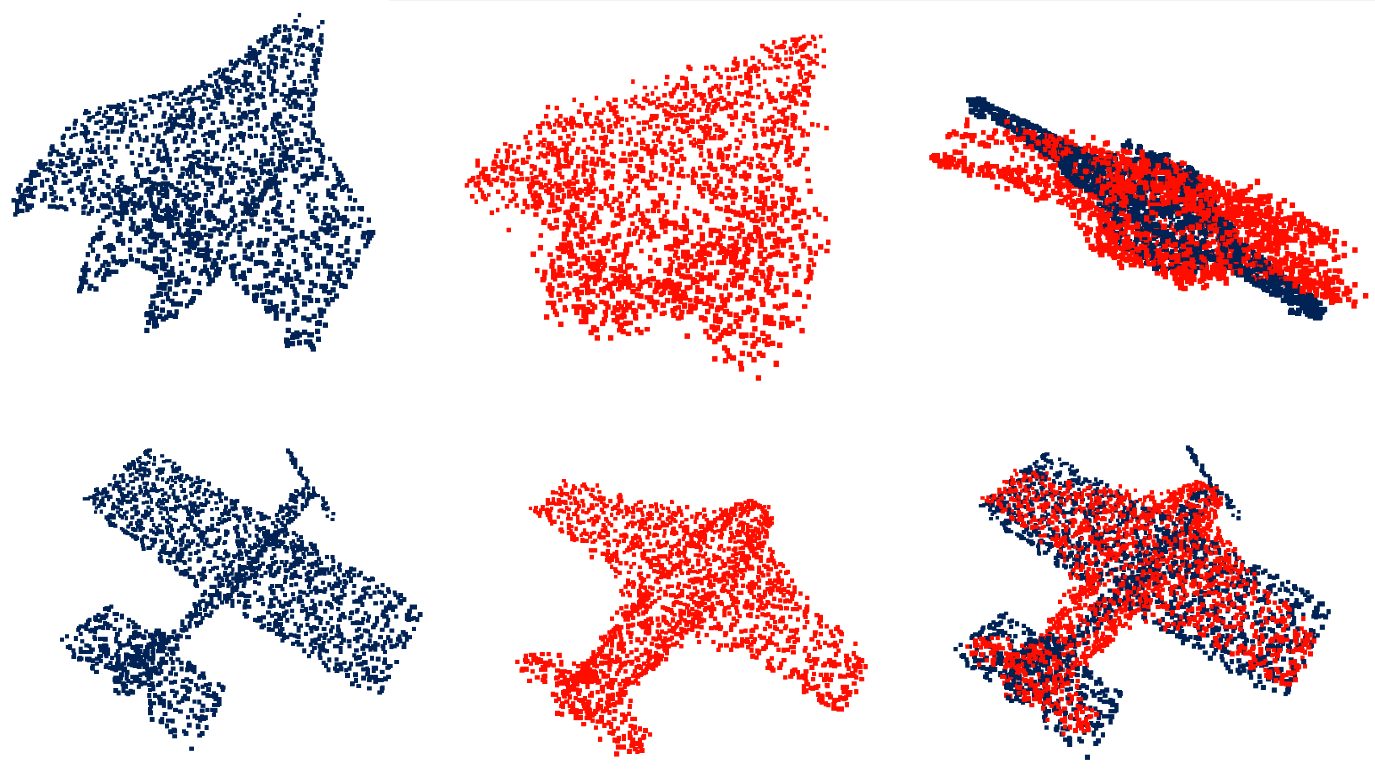}
\end{center}
   \caption{\textbf{DAE failure cases.} The Denoising Autoencoder is notoriously prone to overfitting on the training data. Overfitting increases with the percentage of the incompleteness with which they are trained. Column 1 has complete point clouds from the test set which were fed to the DAE without any masking. Even with a complete cloud as input, the DAE outputs a different point cloud (Column 2), since it essentially collapses to a Nearest Neighbour search against the training data. The difference between input and DAE output is visualized in column 3. Due to lack of large datasets for 3D, this is a huge disadvantage of the plain DAE. We also note in our experiments that a DAE trained with higher percentages of missing data performs poorly on lower percentages of corruption.}
\label{fig:dae}
\end{figure}

\begin{figure}
\begin{center}
\includegraphics[width=1\linewidth]{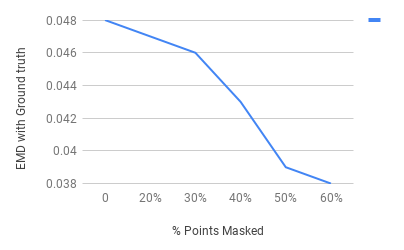}
\end{center}
\caption{ The plot shows the performance of Denoising Autoencoder(trained on point clouds with 60\% masking) with different levels of masking. The plot shows that the DAE performs worse as the masking is reduced. Infact, it gives the worst performance when the ground truth (0\% masking) is given as input.}\label{fig:dae_overfit}
\end{figure}

We observe that the DAE overfit to the specific noise type they are trained on. If we train the DAE with a masking of 60\% and test it on lower levels of masking, it's performance decreases. This can be seen in Figure \ref{fig:dae_overfit}. Taking a closer look at the completion results of the DAE, we find that the DAE has simply learnt a fixed mapping in the training set from masked clouds to the completed point clouds. When exposed to unseen data such as the test set, it simply produces a nearest neighbor from the training set. This can be seen in Figure \ref{fig:dae}.

\subsection{Variation with masking}
\begin{figure*}
\begin{center}
\includegraphics[width=1\linewidth]{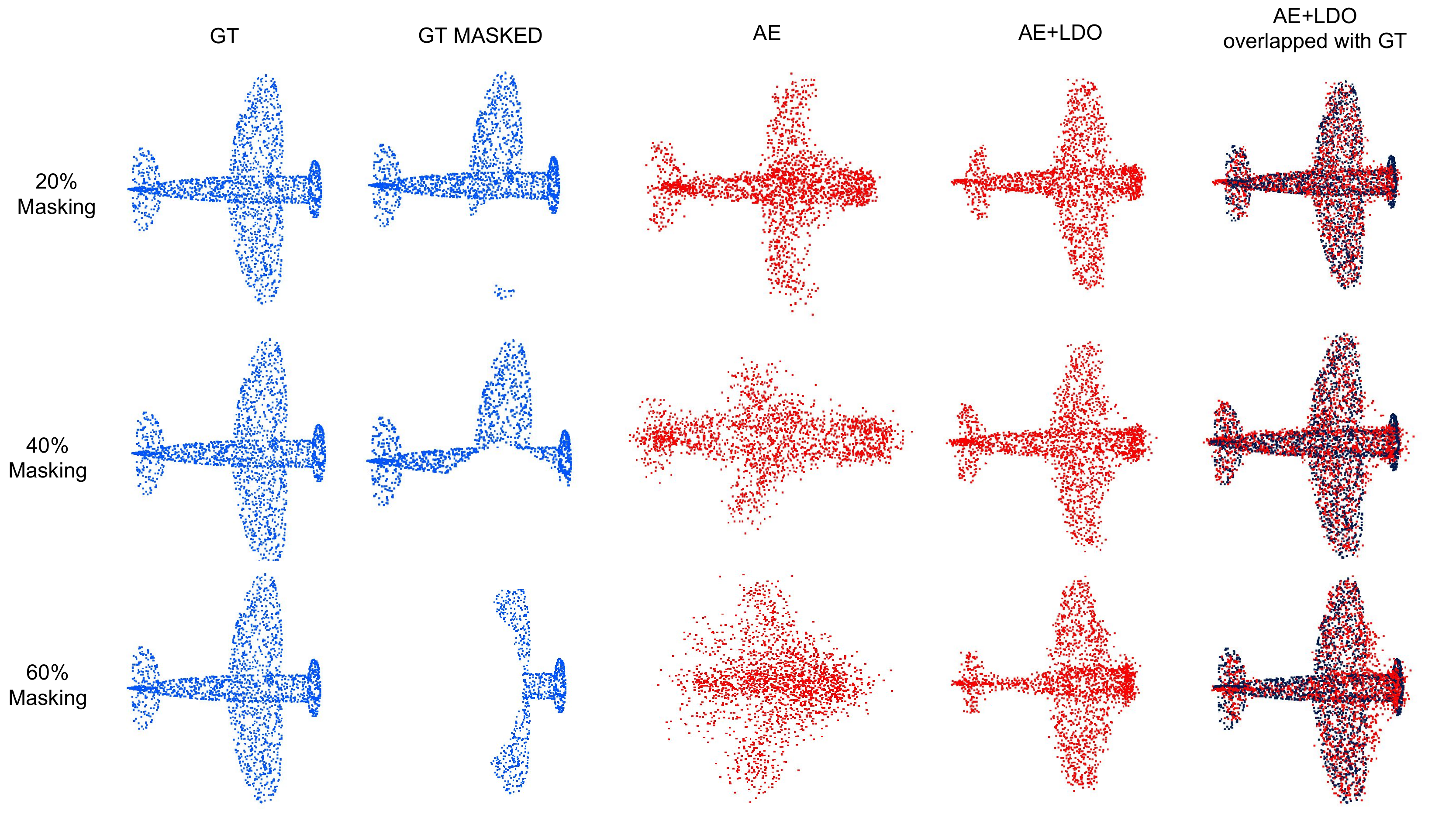}
\end{center}
\caption{ The figure shows the performance of AE and AE+LDO with varying levels of masking (20\%, 40\% and 60\%. The AE in this case is trained specifically on the Airplane dataset.}\label{fig:airplane}
\end{figure*}

Figure \ref{fig:airplane} shows the performance of AE and AE+LDO with varying levels of masking on the Airplane category. The deterioration of performance with increasing levels of masking is clearly visible in the AE but the optimization procedure still manages to generate reasonable looking point clouds. This shows the robustness of our approach to different kinds of point cloud deformations. 

\begin{figure*}
\begin{center}
\includegraphics[width=0.8\linewidth]{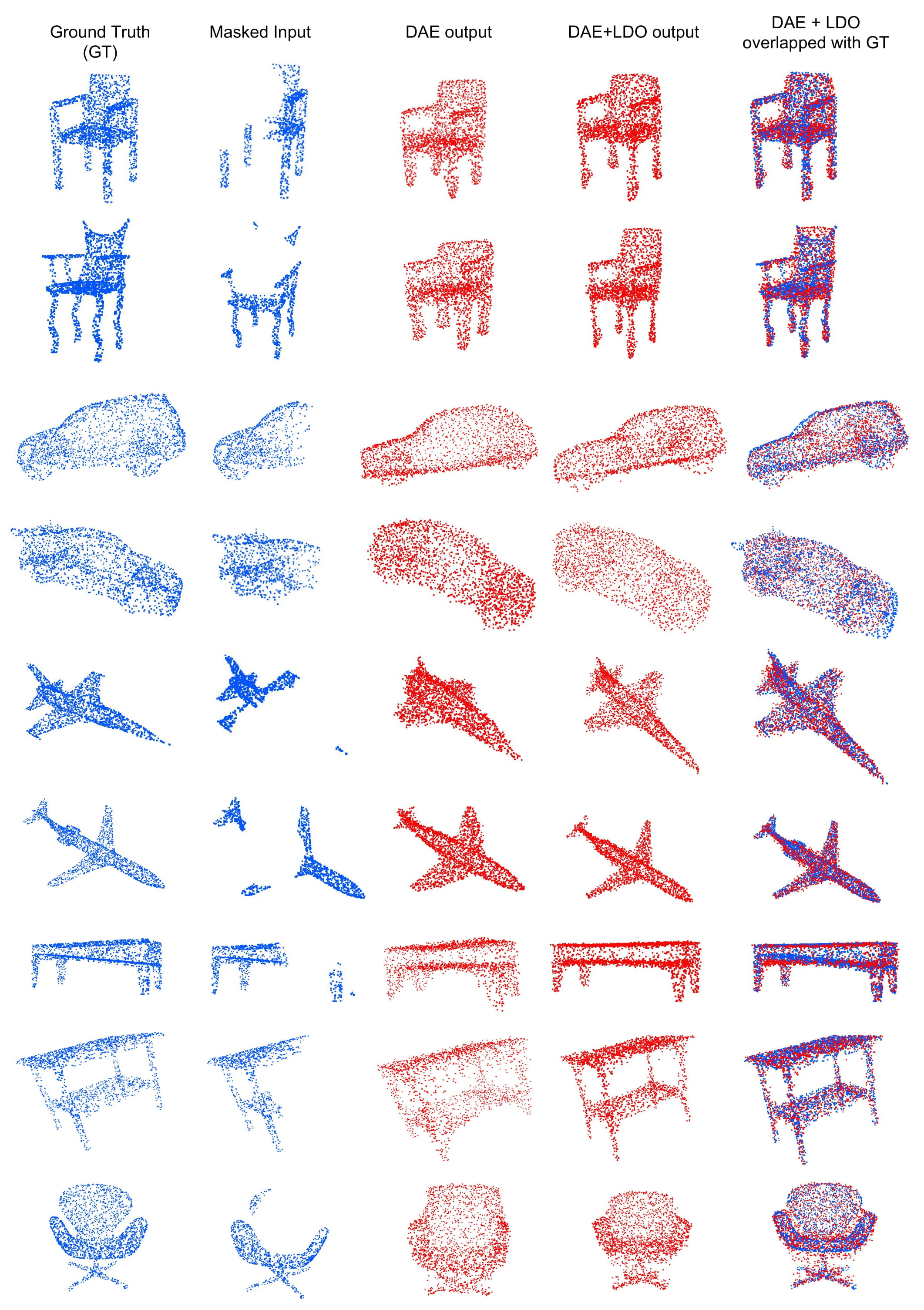}
\end{center}
\caption{ Enlarged version of Fig. 3 results in main paper, showing reconstructions of Multi-Class DAE and Multi-Class DAE+LDO}\label{fig:results_big}
\end{figure*}

\begin{figure*}
\begin{center}
\includegraphics[width=0.9\linewidth]{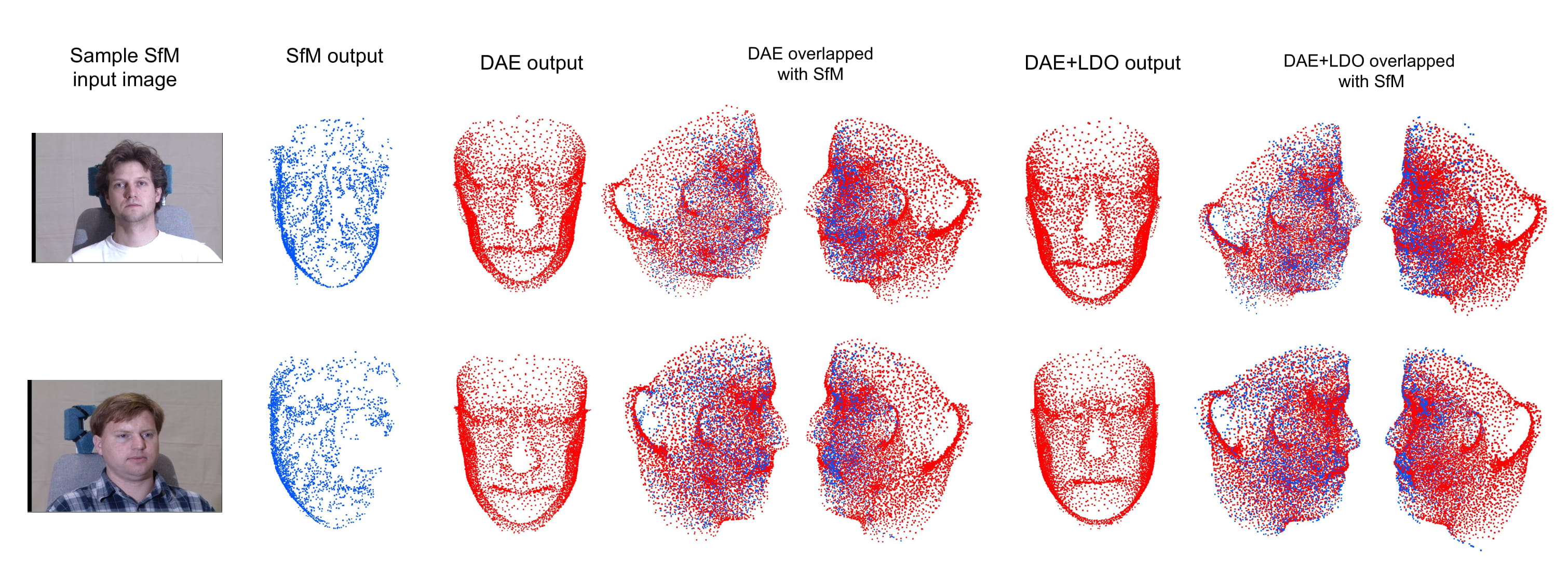}
\end{center}
\caption{ Profile views of reconstructions of the faces shown in  Fig. 7 (SfM data) results in main paper, showing reconstructions of Face DAE and Face DAE+LDO (8k points)}\label{fig:face_sfm}
\end{figure*}

\begin{figure*}
\begin{center}
\includegraphics[width=0.9\linewidth]{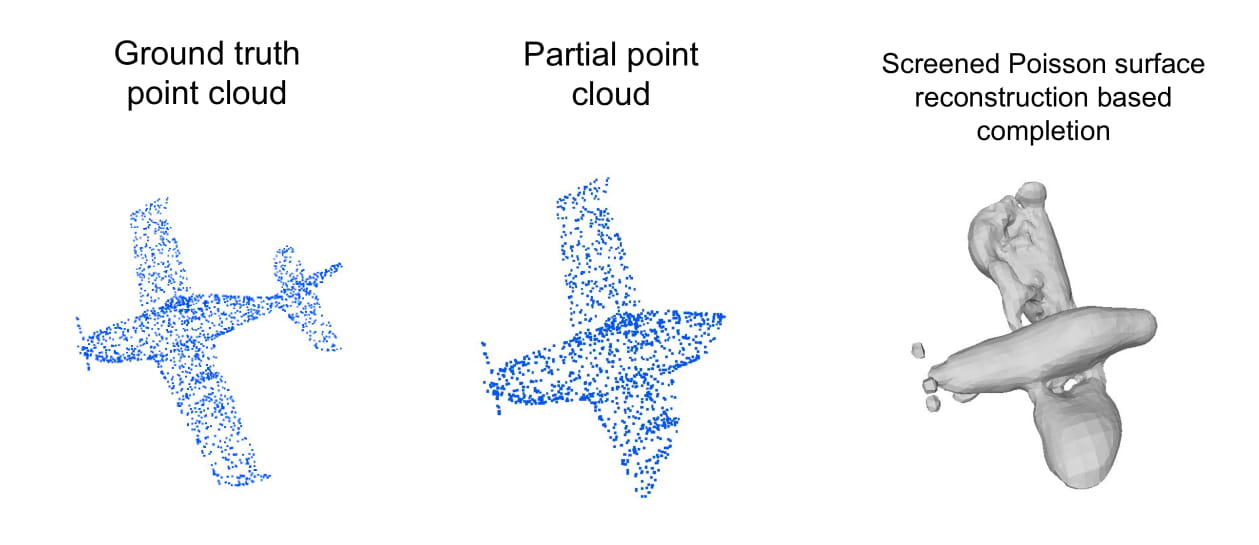}
\end{center}
\caption{Geometric methods such as Screened poisson reconstruction, while effective at removing small holes in surfaces, fail on large missing regions.}\label{fig:screened}
\end{figure*}

\end{document}